\documentclass{article}

\usepackage[final]{corl_2026} 
\usepackage{comment}
\usepackage{float}
\usepackage{colortbl}

\usepackage{enumitem}

\title{Which Hand Was Late? Diagnosis for Robot Failure Recovery in Bimanual Tool Use}

\usepackage{graphicx, xcolor, amsmath, booktabs}

\newcommand{\baptiste}[1]{\textcolor{teal}{(baptiste:\ #1)}}

\author{
  Baptiste Bonin$^{1, 2}$ \quad
  Caro Strickland$^{1,2}$ \quad
  Audrey Durand$^{1,2,3}$ \quad
  \\ \\
  $^{1}$Université Laval \qquad
  $^{2}$Mila -- Quebec AI Institute\qquad
  $^{3}$CIFAR AI Chair
}

\begin{document}
\maketitle


\begin{figure}[H]
\centering
    \includegraphics[width=\linewidth]{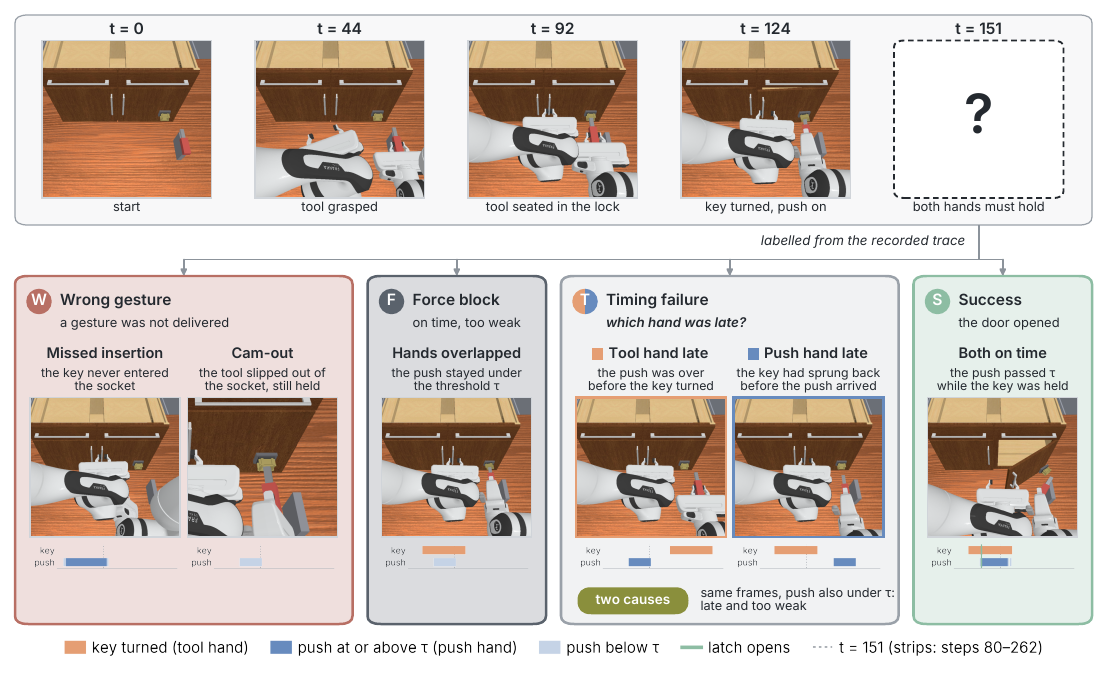}
\vspace{-1.6em}
\caption{\textbf{Which hand was late?} One hand turns a key-tool in the lock while the other pushes the door (top). Each attempt is labelled from its recorded trace (strips): a wrong gesture (W), a push too weak while both hands acted (F), right gestures at different times (T), or success (S). A late attempt whose push was also below the hidden threshold $\tau$ has the same frames and two causes.}
\label{fig:diag}
\end{figure}


\begin{abstract}
Failure diagnosers for robot manipulation are increasingly vision--language models (VLMs), trained and evaluated on single-arm failures produced by perturbing a successful demonstration. Bimanual tool use adds failures of other kinds: a tool that slips out of its socket while still held, two correct gestures made at different times, and attempts that fail for two reasons at once. On a simulated bimanual tool-use task whose failure causes are read from the recorded trace, we measure what each sensing channel lets a diagnoser tell. The tool's slip is read on the tool hand; which hand was late is read on both arms, while a sensor on one hand reads its own hand's delay and at most part of the other's; and whether a late attempt was also pushed too weakly is not identifiable from passive sensing when the tool hand is late. Within two attempts, a repair policy that escalates when the cause is not identifiable repairs as many failures as one told the true cause. A VLM's internal state carries the late hand almost perfectly, along a direction no closer to the one its answer reads than a random one; whether the answer reports it depends on the wording of the question.
\end{abstract}



\section{Introduction}

Imitation learning and recent Vision-Language-Action (VLA) models have enabled impressive advances in robotic manipulation \cite{kim2024openvla, black2025pi} and with them the question of what a robot should do when an attempt fails. Deployed in an open world, a robot will encounter mechanisms it has never interacted with before~\cite{tedrake2024manipulation}, and it will inevitably fail on some of them. Some of what a mechanism requires can only be found out by trying: a robot that has never opened a stiff lock cannot know in advance how much force it requires. What matters is whether it can learn from the attempt. Failures can also occur on familiar tasks when the robot misestimates quantities that are not directly observable, such as the force required to push a locked door open or the timing of the other hand's action. In either case, a failure is useful only if its cause can be identified: to recover from a failure and learn from it, a robot must detect it, attribute it to its cause, and act on that cause.


A growing body of work addresses this problem through visual failure reasoning by asking a vision--language model (VLM). AHA~\cite{duan2025aha} takes an image sequence and returns a natural-language explanation of the failure; RoboFAC~\cite{lu2025RoboFac} evaluates detection, identification and localisation from video; 
ViFailback~\cite{Zeng2026Diagnose} turns the diagnosis into visual corrections for a policy; and ARMOR~\cite{qi2026selfrefining} jointly refines detection and open-ended reasoning. These systems address \textbf{detection}, \emph{did the attempt fail?}, and \textbf{diagnosis}, \emph{why did it fail?} Both are learned from failures of a particular kind: each involves one arm acting on one object, is produced by perturbing a successful demonstration, and has a cause that is visible in the scene.

Bimanual tool use breaks each of these assumptions. Consider unlocking a utility cabinet with a square key: one hand seats the key in the lock and turns it, while the other pushes the door. The attempt can fail because a gesture was wrong (the key was never seated), because the tool slipped out of its socket under torque while still firmly held, because the lock needed more force than was applied, or because both hands did the right thing \emph{at different times}. The last case is not a mistake of either arm but a relation between them. For such a failure, ``why did it fail?'' becomes ``\emph{which hand was late?}'', a question no single-arm taxonomy can ask (Figure \ref{fig:diag}). In this paper, we ask:

\begin{quote}
\emph{What must a failure diagnoser observe to tell which hand caused a bimanual tool-use failure, and
do VLM diagnosers report it when they observe it?}
\end{quote}

To answer this, we need a task in which the cause of each failure is known without relying on an
annotator's reading of the video. We build a bimanual, tool-mediated lock in simulation, inject one fault per attempt with a scripted controller, and read the class of every failure from the recorded trace of the mechanism. We then apply the same probe to one sensing channel at a time, and to the internal representation and the language output of the same VLM. We do not propose a new diagnoser, we instead measure what a diagnoser needs. Our contributions are the following:
%
%
\begin{enumerate}[leftmargin=*]
\item \textbf{A tool-use task with known failure causes} (Sec.~\ref{sec:task}), where the cause of each failure can be recovered from the trace, including failures with multiple simultaneous causes.
\item \textbf{What can be diagnosed, and from where} (Sec.~\ref{sec:where}): the tool's slip is read on the tool hand and which hand was late on both arms, while whether a late push was also too weak is on no passive sensor when the tool hand is late. Sensor placement determines which of these signals are available.
\item \textbf{Diagnosis determines repair} (Sec.~\ref{sec:repair}): failures with two causes are repaired only when both causes are corrected. A policy that asks only whether the hands overlapped, applies the matching correction and, if it fails, both (\emph{escalation}) repairs as many failures within two attempts as one told the true cause.
\item \textbf{The model represents the late hand} (Sec.~\ref{sec:readout}): the VLM’s state encodes which hand was late, even though this information is not read out by its answer.
\end{enumerate}

\section{Related work}

Failure diagnosis with vision--language models is a recent line of work, almost entirely concerned with a single arm and with judging an explanation by its text. We review these diagnosers, how their answers are scored, and the two tools our study borrows: bimanual platforms and representation probes.

\textbf{Failure reasoning with VLMs.} Beyond the systems above, post-trained failure detectors such as I-FailSense~\cite{ifailsense2026} fine-tune a VLM (here PaliGemma-2) specifically for robot failure detection. REFLECT \cite{liu2023reflect} is the closest to our setting in that it uses non-visual modalities, but it treats them as evidence for a scene-level explanation rather than asking which modality carries which cause.

\textbf{How answers are scored.} These systems score an explanation with a classifier head or by its similarity to a reference: ROUGE-L~\cite{lin2004rouge}, embedding similarity or an LLM judge. To our knowledge, none compares the answer with what the model's own state contains, which is what our readout analysis does (Section~\ref{sec:readout}).

\textbf{Bimanual platforms and probes.} Low-cost bimanual platforms~\cite{zhao2023aloha} have made
two-arm tasks easy to use, yet we know of no diagnosis benchmark whose taxonomy includes a relation between two arms. Linear probes on frozen features~\cite{alain2017understanding} separate what a representation \emph{contains} from what a model \emph{reports}. We use the same probe on every sensing channel and on the VLM's own states.

\section{A tool-use task with known failure causes}
\label{sec:task}

We need failures whose cause is known without an annotator (contribution~1). We first describe the scene and its lock (Section~\ref{sec:scene}), then how the cause of each failure is read from the recorded trace, including failures with two causes (Section~\ref{sec:labels}), and finally the probe we apply to every sensing channel and to the VLM (Section~\ref{sec:protocol}).
\begin{figure}[t]
\centering
\includegraphics[width=\linewidth]{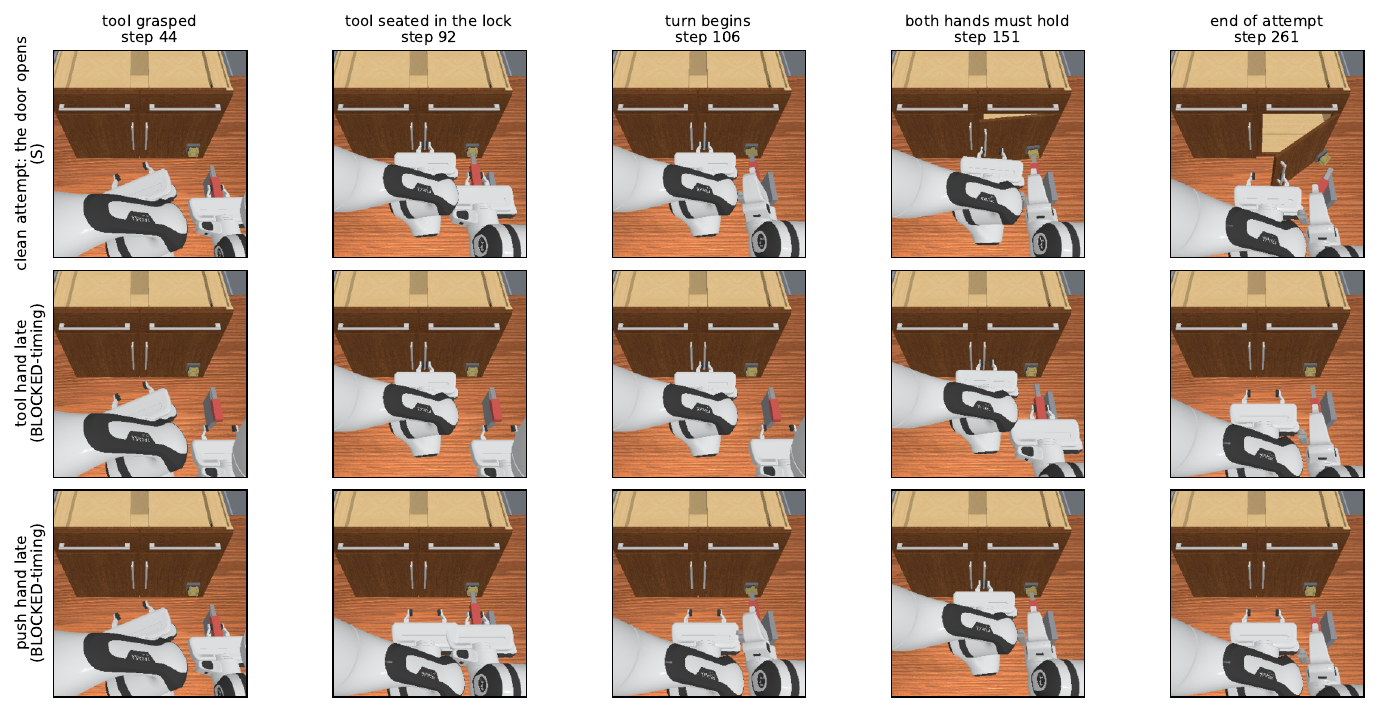}
\caption{The task, three recorded episodes (front camera). One arm fetches the key-tool from its
stand, seats it in the square socket and turns it; the other pushes the door. \textbf{Top:} both hold
their targets at the same time and the door opens. \textbf{Middle:} the tool hand is late; the push
is over before the key turns. \textbf{Bottom:} the push hand is late; the key has sprung back before
the push arrives. In both failures each gesture, taken alone, is correct.}
\label{fig:task}
\end{figure}

\subsection{Scene and mechanism}
\label{sec:scene}

Two Franka Panda arms face a door fitted with a lock (Figure~\ref{fig:task}). One arm fetches a square key-tool from a stand, seats it in the lock's socket and turns it; the other pushes the door. The tool transmits torque through its shape, so the hand never touches the mechanism, and a tool that is not seated deep enough slips out of the socket while still held: a \textbf{cam-out}, which emerges from the contact geometry. The key springs back when released, so both hands must act at once: the door opens only if the key is held turned while the push exceeds a hidden threshold $\tau$, drawn for every attempt and absent from every observation. Details of empirical experiments are in Appendix~\ref{app:task}.

\subsection{Causes read from the trace}
\label{sec:labels}

\begin{figure}[t]
\centering
\includegraphics[width=0.92\linewidth]{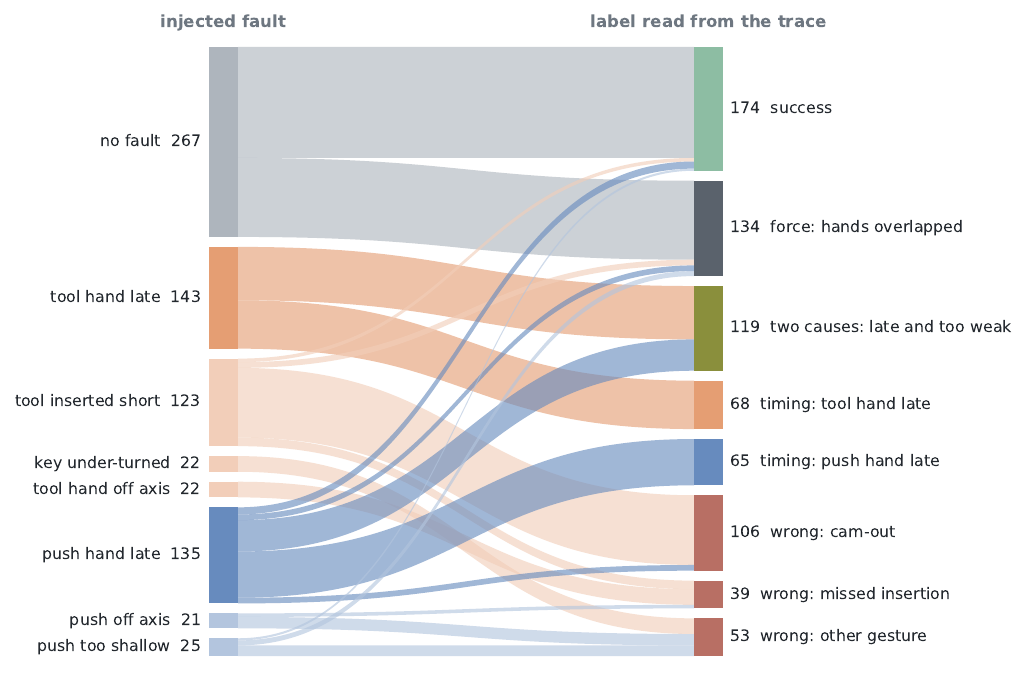}
\caption{Injected fault (left) against the class read from the recorded trace (right), $758$ attempts; the same fault can end in several classes. \emph{Wrong}: a gesture was not delivered. \emph{Timing}: the push reached the hidden threshold $\tau$, but never while the key was held. \emph{Force}: the push was on while the key was held, but stayed under $\tau$. \emph{Two causes}: both at once, the push never on while the key was held and never reaching $\tau$.}
\label{fig:sankey}
\end{figure}

A scripted controller plays a fixed plan for each arm and injects one fault per attempt: the tool inserted short, a hand aimed off axis, the key under-turned, the push too shallow, or one arm's schedule delayed. We label each failure according to the recorded trace (Figure~\ref{fig:sankey}): \textit{wrong} indicates that a gesture was not delivered; \textit{timing} indicates that both gestures were delivered and the push reached the threshold, but never while the key was held as the coordination failure; and \textit{force} indicates that the push was on while the key was held, but never reached the threshold. 

\textbf{Failures with two causes.} An attempt can also fail for two reasons at once: the push was never on while the key was held, \emph{and} it never reached the threshold. Replayed in the same world, re-timing the late hand alone opens the door in $3$ of $119$ attempts and a lower threshold alone in none; doing both opens it in all $119$. These $119$ attempts ($75$ with the tool hand late, $44$ with the push hand late) are neither a timing failure nor a force block, and we keep them as a class of their own.

\textbf{Data and questions.} The campaign has $758$ attempts on $17$ doors: $174$ successes, $198$ \textit{wrong} ($106$ of them cam-outs), $133$ \textit{timing}, $134$ \textit{force} and $119$ failures with two causes; every attempt replays bit for bit. Each question is asked on the population where it makes sense: \emph{detection}, failed against succeeded; \emph{diagnosis}, blocked against \textit{wrong}; \emph{coordination}, timing against force blocks, also split by which hand was late (\emph{tool hand late}, $68$ against $134$, \emph{push hand late} $65$ against $134$); and \emph{cam-out}, a slipped tool against a force block with the tool held ($106$ against $134$). And for failures with two causes, timing against two causes (Section~\ref{sec:twocauses})

\subsection{Probing protocol}
\label{sec:protocol}

We fit the same probe to one input channel at a time: on the tool hand, its wrist torque, the contact between tool and socket, and the joint that turns the key; on the push hand, its wrist force and its contact with the door; joint positions and velocities of both arms (\emph{arm poses}); the commands of both arms; and images from the front and both wrist cameras through a frozen ResNet-50. The \emph{lock angle}, the barrel's own joint and the latch's input, is
reported as a mechanism reading, and the \emph{latch state} (door angle, push over the threshold $\tau$, their conjunction) as an oracle ceiling, never as evidence. Each channel is summarised by its first and last frame in the window and their difference. The probe is an $L_2$-regularised logistic regression, its strength chosen by cross-validation on the training doors; splits are grouped by door in five folds; AUC is computed within each door and pooled by pairs; intervals are $95\%$ bootstraps over doors, and
comparisons between channels are \emph{paired} on the same resampled doors. The \emph{trial window} runs from the moment the key should be turned to the end of the push hold on the nominal schedule; the \emph{approach window} ends before either arm reaches anything and serves as a negative control. Details are in Appendix \ref{app:protocol}.

\section{What can be diagnosed, and from where}
\label{sec:where}

\begin{table}[t]
\centering
\resizebox{0.95\linewidth}{!}{%
\begin{tabular}{@{}l@{\hspace{4pt}}r@{\hspace{4pt}}r@{\hspace{6pt}}cccccc@{}}
\toprule
 & & & & & \multicolumn{3}{c}{coordination} & \\
\cmidrule(lr){6-8}
Input & KiB & GFLOPs & detect & diagnose & tool late & push late & pooled & cam-out \\
\midrule
\textcolor[HTML]{E59E73}{\rule{0.55em}{0.55em}}\,Wrist torque, tool hand & 1.1 & 0.0 & \cellcolor[HTML]{F9EAE1}0.654 & \cellcolor[HTML]{F6DECF}0.746 & \cellcolor[HTML]{EDBB9D}\textbf{1.000} & \cellcolor[HTML]{F6DDCD}0.753 & \cellcolor[HTML]{F2CEB8}0.863 & \cellcolor[HTML]{EEBFA2}0.972 \\
\textcolor[HTML]{E59E73}{\rule{0.55em}{0.55em}}\,Tool--socket contact & 0.7 & 0.0 & \cellcolor[HTML]{F7E1D4}0.719 & \cellcolor[HTML]{F2CEB9}0.858 & \cellcolor[HTML]{EFC2A7}0.949 & \cellcolor[HTML]{FAEBE2}0.646 & \cellcolor[HTML]{F2CFBA}0.853 & \cellcolor[HTML]{EDBDA0}\textbf{0.986} \\
\textcolor[HTML]{E59E73}{\rule{0.55em}{0.55em}}\,Wrist joint 7, tool arm & 0.2 & 0.0 & \cellcolor[HTML]{F8E3D7}0.703 & \cellcolor[HTML]{FCF5F0}0.576 & \cellcolor[HTML]{EDBB9D}\textbf{1.000} & \cellcolor[HTML]{F4D6C4}0.801 & \cellcolor[HTML]{FAEDE5}0.631 & \cellcolor[HTML]{F0C6AD}0.919 \\
\addlinespace[2pt]
\textcolor[HTML]{678BBE}{\rule{0.55em}{0.55em}}\,Wrist force, push hand & 1.1 & 0.0 & \cellcolor[HTML]{F1F4F9}0.568 & \cellcolor[HTML]{E1E8F2}0.640 & 0.398 & \cellcolor[HTML]{9CB3D5}0.966 & \cellcolor[HTML]{D9E2EF}0.678 & \cellcolor[HTML]{FBFCFD}0.519 \\
\textcolor[HTML]{678BBE}{\rule{0.55em}{0.55em}}\,Hand contact, push hand & 0.9 & 0.0 & \cellcolor[HTML]{E1E8F2}0.641 & \cellcolor[HTML]{EDF1F7}0.585 & \cellcolor[HTML]{F1F5F9}0.564 & \cellcolor[HTML]{9BB3D4}0.968 & \cellcolor[HTML]{D5DFED}0.697 & \cellcolor[HTML]{FCFDFE}0.512 \\
\addlinespace[2pt]
\textcolor[HTML]{8DBDA3}{\rule{0.55em}{0.55em}}\,Arm poses, both & 6.5 & 0.0 & \cellcolor[HTML]{BCD8C9}\textbf{0.919} & \cellcolor[HTML]{C3DCCF}0.876 & \cellcolor[HTML]{AFD1BF}\textbf{1.000} & \cellcolor[HTML]{B4D4C2}0.970 & \cellcolor[HTML]{B1D2C0}\textbf{0.986} & \cellcolor[HTML]{B6D5C4}0.959 \\
\textcolor[HTML]{8DBDA3}{\rule{0.55em}{0.55em}}\,Commands, both & 2.9 & 0.0 & \cellcolor[HTML]{CAE0D4}0.833 & \cellcolor[HTML]{C5DED0}0.862 & \cellcolor[HTML]{B5D4C3}0.964 & \cellcolor[HTML]{B4D3C2}\textbf{0.972} & \cellcolor[HTML]{B6D5C4}0.957 & \cellcolor[HTML]{B5D4C3}0.965 \\
\textcolor[HTML]{A88BB8}{\rule{0.55em}{0.55em}}\,Images, 3 cameras & 486.0 & 13.6 & \cellcolor[HTML]{DACEE1}0.805 & \cellcolor[HTML]{CCBBD6}\textbf{0.916} & \cellcolor[HTML]{C2AECD}\textbf{1.000} & \cellcolor[HTML]{C6B3D0}0.968 & \cellcolor[HTML]{C4B1CF}0.982 & \cellcolor[HTML]{CEBED7}0.902 \\
\midrule
\textcolor[HTML]{AEB5BD}{\rule{0.55em}{0.55em}}\,\emph{Lock angle (mechanism)} & 0.4 & 0.0 & \cellcolor[HTML]{D4D8DC}\emph{0.876} & \cellcolor[HTML]{DDE0E3}\emph{0.800} & \cellcolor[HTML]{C6CBD1}\emph{1.000} & \emph{0.334} & \cellcolor[HTML]{E1E4E7}\emph{0.764} & \cellcolor[HTML]{C6CBD1}\emph{1.000} \\
\textcolor[HTML]{AEB5BD}{\rule{0.55em}{0.55em}}\,\emph{Latch state (oracle)} & 0.5 & 0.0 & \cellcolor[HTML]{C7CCD1}\emph{0.997} & \cellcolor[HTML]{E7E9EC}\emph{0.709} & \cellcolor[HTML]{C7CCD1}\emph{0.995} & \cellcolor[HTML]{CBD0D5}\emph{0.956} & \cellcolor[HTML]{C9CED3}\emph{0.974} & \cellcolor[HTML]{E0E3E6}\emph{0.769} \\
\midrule
\textcolor[HTML]{A88BB8}{\rule{0.55em}{0.55em}}\,Qwen3-VL 8B, frozen features & 294 & 3,056 & \cellcolor[HTML]{E9E1ED}0.682 & \cellcolor[HTML]{E4DBE9}0.719 & \cellcolor[HTML]{C2AECD}1.000 & \cellcolor[HTML]{C5B2D0}0.973 & \cellcolor[HTML]{C4B0CF}0.985 & \cellcolor[HTML]{F4F1F6}0.588 \\
\textcolor[HTML]{A88BB8}{\rule{0.55em}{0.55em}}\,Qwen3-VL 8B, prompted answer & 294 & 3,056 & \cellcolor[HTML]{FEFDFE}0.511 & \cellcolor[HTML]{F7F4F8}0.570 & 0.497 & \cellcolor[HTML]{F2EEF5}0.604 & \cellcolor[HTML]{FFFFFF}0.501 & 0.475 \\
\bottomrule
\end{tabular}
}
\vspace{0.4em}
\caption{Input cost per episode and information separated (trial window; within-door AUC). KiB denotes the input size: all samples in the window for numerical channels (a probe reads two channels) and two frames for images; GFLOPs denotes feature-extraction cost. Bold indicates the best robot-accessible probe per column. Lock angle is a mechanism reading and latch state an oracle; neither is evidence. Coordination measures timing relative to force; cam-out measures the corresponding $106/134$ trial separation. The VLM uses the front-camera image pair.}
\label{tab:channels}
\end{table}

This section applies the probe of Section~\ref{sec:protocol} to each sensing channel, one question of Section~\ref{sec:labels} at a time (contribution~2). Every number is a within-door AUC: the probability that the probe ranks an attempt of the first class above one of the second on the same door, so that $0.5$ is chance and $1$ a perfect separation. Table~\ref{tab:channels} gathers them, one row per input and one column per question, next to what each input costs per attempt. Every channel that reads both arms separates a blocked attempt from a wrong gesture (arm poses $0.876$, commands $0.862$, images $0.916$): the wrong gestures are geometric, and the arms display them. What remains is to locate the cause of a blocked attempt: on one arm
(Section~\ref{sec:camout}), between the two arms (Section~\ref{sec:bilateral}), or on no sensor
(Section~\ref{sec:twocauses}).

\subsection{On one arm: the tool's own failure}
\label{sec:camout}

A cam-out is read on the hand that holds the tool: its wrist torque identifies it at $0.972$ $[0.930, 1.000]$, the contact between tool and socket at $0.986$ $[0.959, 1.000$], and the wrist joint that turns the key at $0.919$ $[0.818, 0.997]$, while the push hand's force and contact stay at chance ($0.519$, $0.512$); channels on both arms, which include the tool arm, read it too (arm poses $0.959$, commands $0.965$). Images from three cameras identify it at $0.902$ $[0.808, 0.966]$; the front camera alone, the one the VLM is given, at $0.768$ $[0.559, 0.908]$. A tool failure is thus localised on the hand that holds the tool.

\subsection{Between the arms: which hand was late}
\label{sec:bilateral}

\begin{figure}[t]
\centering
\includegraphics[width=\linewidth]{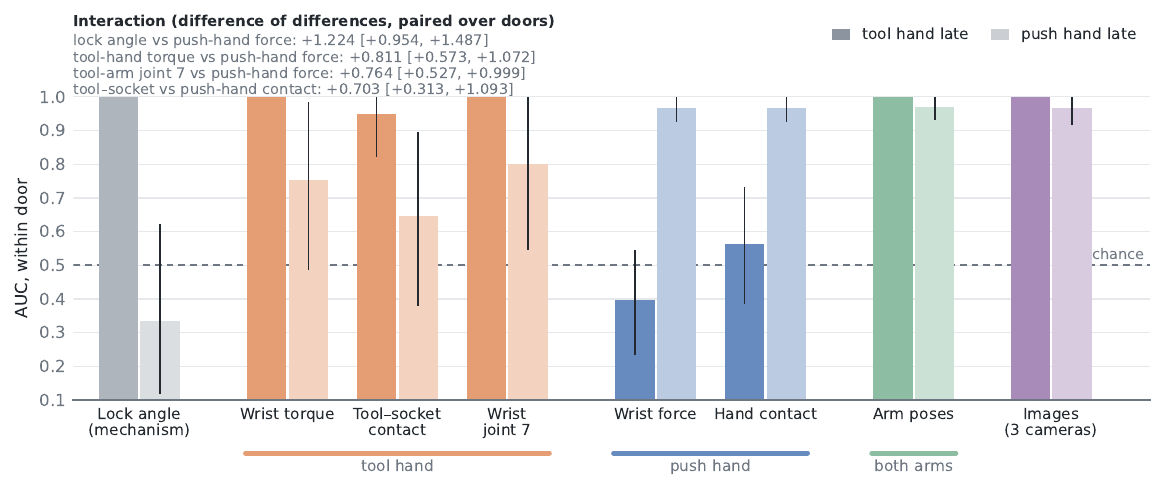}
\vspace{-2em}
\caption{Channels across the two halves of the coordination failure (trial window; within-door AUC with $95\%$ intervals over doors; timing relative to force). Colour indicates sensor location; solid and light lines denote \emph{tool-hand late} and \emph{push-hand late}, respectively; grey shows lock angle. One-hand channels primarily capture their own hand’s delay, whereas two-arm channels capture both. Top: interaction effects (difference-in-differences) across paired door resamples.} 
\label{fig:crossed}
\end{figure}

Sensors on one hand read that hand's delay, and the other's only in part or not at all
(Figure~\ref{fig:crossed}). The tool hand's wrist torque identifies \emph{tool hand late} at $1.000$ and \emph{push hand late} at $0.753$ $[0.485, 0.985]$, the joint that turns the key $1.000$ and $0.801$ $[0.546, 1.000]$; the push hand's wrist force does the reverse, $0.398$ $[0.234, 0.546]$ and $0.966$ $[0.926, 1.000]$, and the contacts follow ($0.949$/$0.646$ against $0.564$/$0.968$). Tested as an interaction (the difference between two channels on one half minus the same difference on the other, over the same resampled doors), the dissociation is $+0.811$ $[+0.573, +1.072]$ for tool-hand torque against push-hand force, $+0.764$ $[+0.527, +0.999]$ for the wrist joint and $+0.703$ $[+0.313, +1.093]$ for the contacts; the lock angle, a reading of the mechanism, gives the sharpest ($+1.224$ $[+0.954, +1.487]$). The asymmetry matters for repair: a late tool hand leaves the push unchanged, the same force on every such attempt of a door, so the push hand's sensors take it for a force block, and a repair based on them would push harder.

\textbf{Channels on both arms read both halves.} Arm poses reach $1.000$ and $0.970$, commands $0.964$ and $0.972$, images $1.000$ and $0.968$.
Asked only whether the hands were out of sync, whichever was late, arm poses keep $0.986$, while one-hand channels drop to $0.631$--$0.863$. The cheapest sensors suffice if they cover both arms: arm poses are $108$ numbers, against $3{,}056$\,GFLOPs for one call to Qwen3-VL 8B (Table~\ref{tab:channels}). 

\textbf{How robust the pattern is} (numbers in Appendix~\ref{app:protocol}). It holds on the full time series of the window and under sensor noise of $0.2$ times each dimension's spread, where all three interactions stay above $+0.68$; before contact, in the approach window, every channel is at chance or barely above it ($\le 0.54$). It is tied to the delay schedule, however: when the two arms start at random times, arm poses, commands and images no longer read the late hand (all $\le 0.65$), although the arms' relative onset alone separates the halves ($1.000$ and $0.990$). 

\subsection{On no sensor: failures with two causes}
\label{sec:twocauses}

A failure with two causes and a timing failure look alike: in both, the push was never on while the key was held. However, they differ in one fact, whether the push would have reached the threshold, and the threshold is in no observation. How well can a diagnoser do at best? On the attempts in which the hands never overlapped ($226$: $107$ timing failures, $119$ with two causes), we computed, before fitting any probe, the best AUC any passive reading of the attempt can reach: rank attempts by the probability, under the prior on $\tau$, that the push force they delivered exceeds $\tau$. When the tool hand is late this ceiling is exactly $0.500$: within a door the push hand follows the same plan on every such attempt and delivers the same force, so whether it beats the threshold is decided by the threshold alone. Over all $226$ attempts, whichever hand was late, the ceiling is $0.532$, and $0.509$ when the push hand is late.

The probes stay at these ceilings. When the tool hand is late, no channel exceeds $0.554$ (lock angle, wrist torque and joint, push-hand force and contact, arm poses, commands) while the latch-state oracle, which contains the threshold $\tau$, reaches $1.000$; over all $226$, channels lie between $0.448$ and $0.540$ (oracle $0.851$). When the push hand is late, on $83$ attempts ($10$ doors with both classes), two channels exceed the ceiling: the push hand's contact ($0.731$ $[0.546, 0.848]$) and the commands ($0.616$ $[0.526, 0.734]$); redrawing $\tau$ from its prior and refitting the same probe gives AUCs this high in $0.5\%$ and $2.7\%$ of the draws ($47\%$ for the push hand's force). The ceiling assumes that no recorded signal depends on $\tau$ before the latch opens, since $\tau$ only decides whether it opens; for these two channels the assumption may fail, although with $24$ numeric cells tested the two values remain compatible with chance once the selection is accounted for ($p = 0.12$ and $0.65$). Images from three cameras stay at the ceiling too: $0.515$ pooled, $0.415$ when the tool hand is late and $0.553$ $[0.304, 0.746]$ when the push hand is late.

\section{Diagnosis decides the repair}
\label{sec:repair}

To test whether diagnosis decides the repair (contribution~3), we replay every blocked attempt in the same world under each correction (Table~\ref{tab:repair} in Appendix~\ref{app:task}). A timing failure is repaired by re-timing ($127$ of $133$), a force block by pushing harder ($81$ of $134$ at three times the depth), and a failure with two causes only by both ($81$ of $119$; re-timing alone $3$, pushing harder alone $0$). A diagnoser that names a single cause repairs $211$ of the $386$ attempts within two tries. Asking instead only whether the hands overlapped (an arm-pose probe answers at $0.992$ $[0.979, 1.000]$), applying the matching correction, and escalating to both if it fails repairs $297$ ($+0.223$ $[+0.169, +0.268]$ over one label, paired by door), as many as an oracle told the true cause or as all corrections at once, with $323$ pushes harder instead of $386$. The oracle repairs more at the first attempt: $289$ against $200$ ($+0.231$ $[+0.180, +0.284]$). When the cause is unidentifiable, the right diagnosis says so and repairs for both.

\section{The model knows the late hand but does not say it}
\label{sec:readout}

\begin{figure}[t]
\centering
\includegraphics[width=\linewidth]{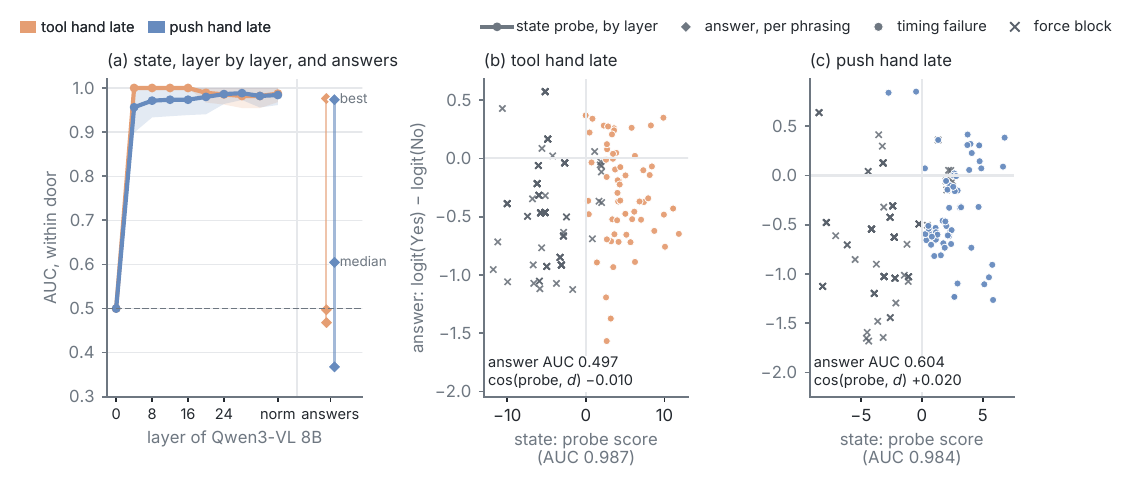}
\vspace{-2em}
\caption{What Qwen3-VL 8B holds and what it answers (front-camera frames, corrected labels). \textbf{(a)} Within-door AUC of a linear probe on the residual stream at the position that predicts the answer, by layer (shaded: 95\% intervals over doors), and the AUC of the answer under each of the three phrasings (diamonds). \textbf{(b, c)} One point per attempt: the probe's out-of-fold score on the final-norm input, against the answer's $\log P(\text{Yes}) - \log P(\text{No})$ under the phrasing whose state the probe reads.}
\label{fig:readout}
\end{figure}

\textbf{How the VLM is asked, and scored.} To test whether the model represents the late hand and says it (contribution~4), we give it the two front-camera frames (first and last of the trial window) and a yes/no question in three phrasings per half (e.g., \emph{``Did the hand on the tool act too late -- turning the tool only after the other hand had
already finished pushing?''}). The score is $\log P(\text{Yes}) - \log P(\text{No})$ at the first answer token, so no generated text is parsed.

\textbf{The answer depends on the wording.} On Qwen3-VL 8B~\cite{qwen3vl2025}, the answers have medians of $0.497$ (\emph{tool hand late}) and $0.604$ (\emph{push hand late}) over the three phrasings, which range from $0.468$ to $0.976$ $[0.935, 1.000]$ and from $0.368$ to $0.973$ $[0.941, 1.000]$ (Figure~\ref{fig:readout}a, diamonds): one phrasing reads both halves almost perfectly, another is at or below chance. On cam-out the answers are at chance ($0.475$). Asked a question that the images answer directly, whether the door is open in the second frame, 
the same model on the same frames scores $0.865$ (over the six doors that have an open attempt, hence the wide interval $[0.334, 1.000]$).

\textbf{The state carries the late hand.} At the answer position, a linear probe on the model's residual stream reads both halves at $0.956$ or more at every layer from layer 4 on (Figure~\ref{fig:readout}a, lines). At the input of the final normalisation, it reads them at $0.987$ $[0.968, 1.000]$ and $0.984$ $[0.961, 1.000]$. The answer reads that same state along one direction, $d = \gamma \odot (W_U[\text{Yes}] - W_U[\text{No}])$, where $W_U$ holds the output embeddings of the answer tokens and $\gamma$ is the gain of the final normalisation. Projecting the state on $d$ reproduces the answers exactly (with the model's bfloat16 rounding). The cosine between $d$ and the probe's direction is $-0.010$ and $+0.020$, as small as between two random directions ($95$th percentile $0.031$ in $4{,}096$ dimensions). Figure~\ref{fig:readout}b--c shows it attempt by attempt: along the probe's score, late attempts and force blocks fall mostly on either side of zero; along the answer's score, the two classes cover the same range.

\textbf{A reference lifts the answer.} One successful attempt on the same door, shown first as a reference, lifts the answers to $0.971$ $[0.921, 1.000]$ on \emph{tool hand late} and $0.793$ $[0.644, 0.924]$ on \emph{push hand late} (medians over phrasings). The probe on the state reads \emph{tool hand late} as before ($0.991$), and $d$ stays as far from the probe's direction as a random direction (cosines $0.011$ and $0.027$): the reference moves the answer without aligning what the answer reads with what the state carries. Nor does the answer, now right, rely on the probe's direction: removing that direction from the state at the layer where the probe reads best (layer 4 for \emph{tool hand late}) changes the answer's AUC by $+0.002$ $[{0.000, +0.006}]$, and by $+0.009$ $[{-0.015, +0.045}]$ on \emph{push hand late}.

\section{Conclusion}

Recovery needs the failure's cause, and in bimanual tool use the cause can be a relation between the hands. On a simulated key-and-lock task whose failure causes are read from the recorded trace, we measured what each sensing channel, and a VLM, lets a diagnoser tell. A tool's own failure is read on the hand that holds it; which hand was late, on both arms, from a few kilobytes of arm poses. When the tool hand is late, whether the push was also too weak is on no passive sensor; a diagnoser that says so and repairs for both causes repairs as many failures as one told the true cause. A VLM holds the late hand in its state, but its answer reads another direction and depends on the wording. These results come from a scripted controller in simulation, and the bilateral pattern holds only under its delay schedule. Next, we will test it with learned policies and real robots, whose delays follow no schedule, on more tasks and tools; study active diagnosis, acting to reveal causes no passive sensor holds; and check VLM answers against the model's state, so that what a diagnoser holds reaches the repair.

\clearpage



\bibliography{example}  
\clearpage
\appendix

\section{Task details}
\label{app:task}

\textbf{Simulation.} ManiSkill3~\cite{taomaniskill3} on the CPU physics backend, two Franka Panda arms in joint-position control, cabinets from PartNet-Mobility~\cite{xiang2020sapien}. An attempt lasts $262$ control steps.

\textbf{Lock and tool.} The lock is a flush barrel with a $16$\,mm square socket, $20$\,mm deep, whose entry is chamfered at $45^\circ$ over $4$\,mm. The key-tool has a $26$\,mm square handle and a $14$\,mm square drive (friction coefficient $0.3$); the nominal insertion is $16$\,mm. Torque passes from hand to tool and from tool to barrel by form closure. A tool pushed past the chamfer bears on straight walls and drives the barrel; a tool left in the chamfer bears on inclined faces whose reaction pushes it out, and it spins in place (cam-out).

\textbf{Latch.} The key is spring-returned. The latch is checked once per control step and releases when the barrel is turned by at least $0.6$\,rad and the push on the door is at least $\tau$ at that step; $\tau$ is drawn log-uniformly in $[20, 700]$\,N for every attempt. Once released, the latch stays open and a spring swings the door, so the outcome does not depend on how long the arm keeps pushing.

\textbf{Controller.} Both arms follow scripted, open-loop joint-space plans. The tool arm approaches, grasps the tool on its stand, lifts, transfers, aligns, inserts, turns ($75^\circ$) and holds; the push arm grasps the handle and pushes, lengthening its approach so that both arms move concurrently. A scripted controller, rather than a learned policy, makes each fault a controlled variable.

\textbf{Faults.} An attempt is fault-free with probability $0.35$; otherwise one fault is drawn: tool inserted $8$--$14$\,mm short; tool hand off axis by $4$--$9$ barrel radii; key under-turned (only $5$--$45\%$ of the nominal turn); push off axis by $0.55$--$1.1$ door widths; push $2$--$4.8$\,cm too shallow; tool hand late by $74$--$88$ steps; push hand late by $64$--$96$ steps. A delay shifts every segment of that arm after its approach; the delay ranges are set so that, on the nominal schedule, the two holds do not overlap.

\textbf{Classes.} From the recorded trace, with a step \emph{keyed} if the barrel is turned by at least $0.6$\,rad and in \emph{contact} if the push is at least $5$\,N (the controller's own delivered-push floor): \textit{success} if the latch opened; \textit{wrong} if the barrel never reached $0.6$\,rad or the push never
reached $5$\,N; \textit{timing} if the push reached $\tau$ (necessarily without the key held); \textit{force} if at least one step was keyed and in contact, and the push stayed under $\tau$ whenever the key was held; \emph{two causes} if no step was keyed and in contact, and the push stayed under $\tau$. The largest keyed push of a two-cause attempt is $0.04$\,N and the smallest of a force block $14.1$\,N, so any contact threshold in between gives the same split ($134$ force, $119$ two causes); $26$ timing failures also satisfy the force definition and keep their timing class. Which hand was late is the delayed arm; on all $133$ timing failures it agrees with the order of the two intervals in the trace. The tool's outcome is read from its pose in the socket: engaged from $2.5$\,mm of depth within $8$\,mm of the axis; cam-out if, before the commanded withdrawal, it twists more than $0.30$\,rad relative to the barrel or leaves the socket; missed if never engaged. Ten delayed attempts succeeded: in nine the key did not spring back after the tool arm withdrew, and in one the push hand's grasp of the handle alone exceeded $\tau$.

\textbf{Campaign and replay.} We drew $1{,}000$ seeds over the $23$ doors; $17$ qualify for the tool gesture (for four no stand position is reachable, for two the tool does not enter the socket), and seeds that fall on the other six are skipped rather than replaced, so that each seed keeps its door and threshold. Every one of the $758$ attempts was replayed from its recorded actions in a separate process: labels, joint positions of both arms, barrel angle, push force, tool-hand torque, tool pose and the rendered images at shared steps were identical, with a maximum deviation of exactly zero.

\textbf{Repair replays.} Every blocked attempt was replayed in the same world under each correction: re-timing (the late hand's delay removed), pushing harder ($0.066$\,m commanded depth, three times the nominal), both, and, as a counterfactual, a lower threshold ($\tau = 0.5$ times the push force the attempt reached). Ten unmodified replays reproduced the recorded attempts exactly.

\begin{table}[t]
\centering
\scriptsize
\begin{tabular}{@{}lcccc@{}}
\toprule
 & \multicolumn{2}{c}{repaired (of 386)} & two causes & attempts \\
\cmidrule(lr){2-3}
Repair policy & 1st attempt & $\le$ 2 attempts & repaired (of 119) & pushed harder \\
\midrule
One label, one correction at a time & 206 & 211 & 3 & 261 \\
Escalate when the cause is not identifiable & 200 & 297 & 81 & 323 \\
Oracle: true cause & 289 & 297 & 81 & 312 \\
Every correction at once (no diagnosis) & 297 & 297 & 81 & 386 \\
\bottomrule
\end{tabular}
\vspace{0.6em}
\caption{Repairing the $386$ blocked attempts, each replayed in the same world, at most twice. Corrections: re-time the late hand, push harder, or both. \emph{One label} chooses between timing and force; \emph{escalate} only asks whether the hands overlapped, applies the matching correction, then both.}
\label{tab:repair}
\end{table}

\section{Protocol details}
\label{app:protocol}

\textbf{Windows.} The trial window covers steps $106$--$152$ of $262$; the approach window steps $0$--$26$, the last step shared by both arms' schedules before either reaches anything (the image at step $25$ is rendered by replay). The VLM sees the front-camera frames at steps $106$ and $151$.

\textbf{Intervals.} AUCs are computed within each door and pooled by pairs; intervals use $10{,}000$ door resamples, and paired differences and interactions recompute every AUC on the same resample. Interactions and paired differences are reported as bootstrap means.

\textbf{Ceiling for failures with two causes.} On the attempts where no step was keyed and in contact, each attempt is scored by the probability, under the log-uniform prior on $\tau$, that the largest push it delivered exceeds $\tau$ ($10{,}000$ draws per attempt); the within-door AUC of this score against the class is the best a passive reading can reach. It was computed and committed before any probe on that population.

\textbf{Readout geometry.} The state at the answer position is normalised by its root mean square before the probe. Random-direction baselines for the cosine are $10{,}000$ pairs of random unit vectors in $4{,}096$ dimensions (mean $|\cos|$ $0.013$, $95$th percentile $0.031$).

\textbf{Robustness of the bilateral pattern.} Full time series of the trial window: lock angle against push-hand force $+1.305$ $[+1.062, +1.533]$, tool-hand torque against push-hand force $+0.736$ $[+0.510, +1.021]$; with the full series the tool hand's torque also reads \emph{push hand late} ($0.846$ $[0.649, 0.981]$), while the push hand's force reads \emph{tool hand late} at $0.384$ (asymmetry $-0.457$ $[-0.637, -0.257]$). Sensor noise ($0.2\times$ spread, five draws): $+1.116$ $[+0.898, +1.307]$, $+0.681$ $[+0.462, +0.874]$ and $+0.764$ $[+0.545, +0.962]$ for the two contacts; no interaction's interval reaches $0$ up to that level. Approach window: every channel $0.500$ on \emph{tool hand late}; on \emph{push hand late}, arm poses $0.526$ $[0.506, 0.548]$ and the front camera $0.539$ $[0.507, 0.576]$. Randomised onsets (full attempt, temporal probe): arm poses $0.509$ and $0.498$, commands $0.554$ and $0.646$, images (first and last frame) $0.629$ and $0.577$; relative onset $1.000$ and $0.990$; lock angle against push-hand force $-0.347$ $[-0.550, -0.152]$.

\textbf{Second campaign (randomised onsets).} Each arm idles at the start of its plan for $c + a$ (tool) or $c + b$ (push) steps, $c \sim U[0, 5A]$ shared and $a, b \sim U[0, A]$, $A = 256$, with no gesture fault; $400$ attempts on the same $17$ doors ($54$ and $52$ timing failures with the tool and the push hand late); of its $189$ force blocks, $126$ are failures with two causes. Probes read the full attempt. The pre-registered pilot gate on single-arm clock AUCs failed on eight positives, and the campaign was run as a recorded deviation. The pre-registered check (the lower $95\%$ bound of a single-arm clock AUC above $0.65$) is not triggered: the largest, the push arm's start on \emph{push hand late}, is $0.636$ $[0.514, 0.717]$, against $0.542$ expected from the design. Temporal probes of the five one-hand channels and the oracle were run on a cluster (a platform copy; the main campaign's copy differed by at most $0.038$, near chance).

\end{document}